\documentclass{article} 
\usepackage{iclr2021_conference,times}

\iclrfinalcopy

\usepackage{amsmath,amsfonts,bm}

\def\eqref#1{equation~\ref{#1}}

\def\1{\bm{1}}

\DeclareMathAlphabet{\mathsfit}{\encodingdefault}{\sfdefault}{m}{sl}
\SetMathAlphabet{\mathsfit}{bold}{\encodingdefault}{\sfdefault}{bx}{n}

\usepackage{hyperref}
\usepackage{url}

\usepackage{algorithmic}
\usepackage{graphicx}
\usepackage{grffile}
\usepackage{textcomp}
\usepackage{xcolor}
\usepackage{float}
\def\BibTeX{{\rm B\kern-.05em{\sc i\kern-.025em b}\kern-.08em
    T\kern-.1667em\lower.7ex\hbox{E}\kern-.125emX}}
\usepackage{amssymb}
\usepackage{amsthm}

\usepackage{tablefootnote} 
\usepackage{multirow}
\usepackage{multicol}
\usepackage[linesnumbered,ruled,vlined]{algorithm2e}
\usepackage{enumerate}
\usepackage{xspace}
\usepackage[center]{subfigure} 
\usepackage[shortlabels]{enumitem}
\usepackage{booktabs}
\setlist[enumerate]{nosep}
\usepackage{flushend}
\usepackage{mathrsfs}
\usepackage[mathscr]{euscript} 
\usepackage{mdwlist}
\usepackage{comment}

\newtheorem{lemma}{Lemma}

\newcommand{\method}{\textsc{AUBER}\xspace}

\newcommand{\blue}[1]{{\color{black} #1}}

\newcommand{\hide}[1]{}

\title{\method: Automated BERT Regularization}

\author{Hyun Dong Lee\thanks{These authors have contributed equally to the work} \\ Columbia University \\ \texttt{hl2787@columbia.edu} \And
Seongmin Lee\footnotemark[\value{footnote}] \\ Seoul National University \\ \texttt{ligi214@snu.ac.kr} \And
U Kang\thanks{Corresponding author} \\ Seoul National University \\ \texttt{ukang@snu.ac.kr}}

\begin{document}

\maketitle

\begin{abstract}
How can we effectively regularize BERT?
Although BERT proves its effectiveness in various downstream natural language processing tasks,
it often overfits when there are only a small number of training instances.
A promising direction to regularize BERT is based on pruning its attention heads based on a proxy score for head importance.
However, heuristic-based methods are usually suboptimal since they predetermine the order by which attention heads are pruned.
In order to overcome such a limitation, we propose \method,
an effective regularization method that leverages reinforcement learning to automatically prune attention heads from BERT.
Instead of depending on heuristics or rule-based policies, \method learns a pruning policy that determines which attention heads should or should not be pruned for regularization.
Experimental results show that \method outperforms existing pruning methods by achieving up to $10\%$ better accuracy.
In addition, our ablation study empirically demonstrates the effectiveness of our design choices for \method. 
\end{abstract}

\section{Introduction}
\label{sec:intro}
How can we effectively regularize BERT (\cite{BERT})? In natural language processing, it has been observed
that generalization could be greatly improved by fine-tuning a large-scale language model
pre-trained on a large unlabeled corpus. In particular, BERT demonstrated such an
effectiveness on a wide range of downstream natural language processing tasks including question
answering and language inference.

Despite its recent success and wide adoption, fine-tuning BERT on a downstream task is prone to overfitting due to over-parameterization; BERT-base has $110$M parameters and BERT-large has $340$M parameters.
This problem worsens when there are only a small number of training instances available.
Some observations report that fine-tuning sometimes fails when a target dataset has fewer than 10,000 training instances (\cite{BERT,STILTs}).

To mitigate this critical issue, multiple studies attempt to regularize BERT by pruning parameters or using dropout to decrease its model complexity (\cite{SixteenHeads,SpecializedHeads,MixOut}).
Among these approaches, we regularize BERT by pruning attention heads since pruning yields simple and explainable results and it can be used along with other regularization methods.
In order to avoid combinatorial search,
whose computational complexity grows exponentially with the number of heads,
the existing methods measure the importance of each attention head based on heuristics such as an approximation of sensitivity of BERT to pruning a specific attention head.
However, these approaches are based on hand-crafted heuristics that are not guaranteed to be directly related to the model performance, and therefore, would result in a suboptimal performance.

\begin{figure*} [t]
	\centering
	\includegraphics[width=0.97\linewidth]{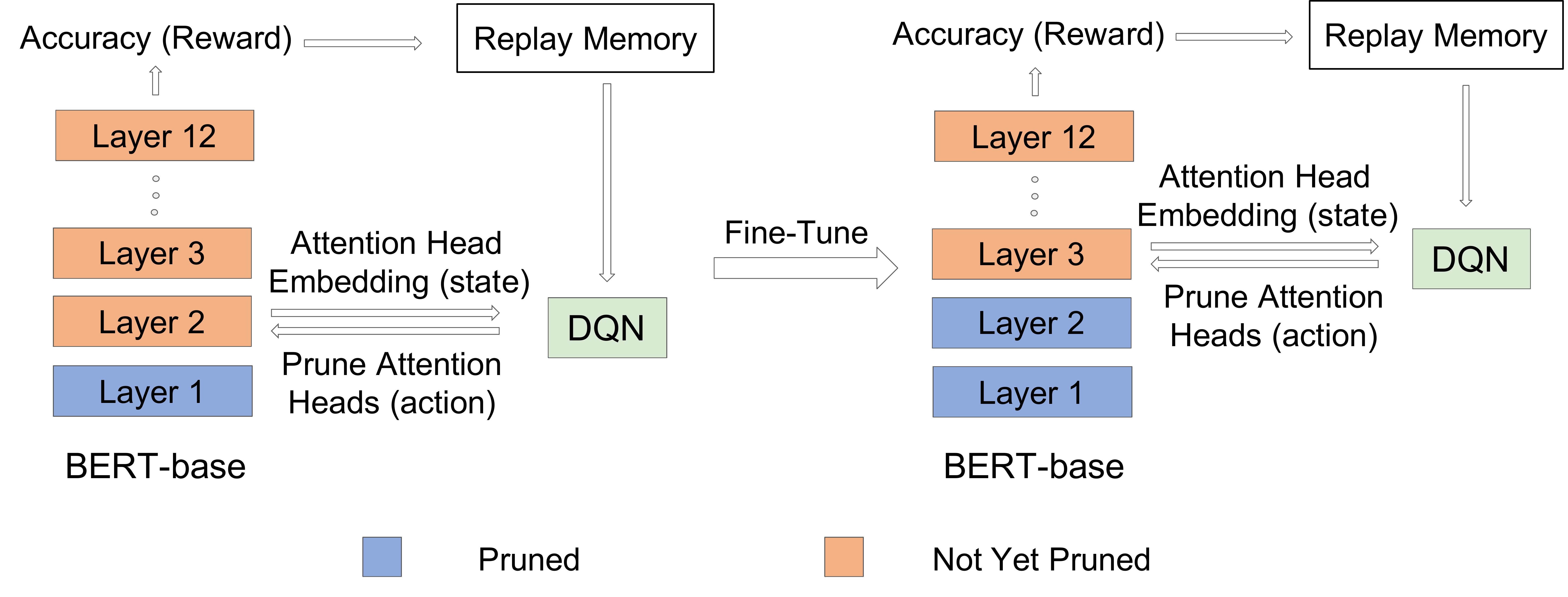}
	\caption{An overview of \method transitioning from $Layer\ 2$ to $Layer\ 3$ of BERT-base. \label{fig:auber}}
\end{figure*}

In this paper, we propose \method, an effective method for regularizing BERT.
\method overcomes the limitation of past attempts to prune attention heads from BERT by leveraging reinforcement learning.
When pruning attention heads from BERT, our method automates this process by learning policies rather than relying on a predetermined rule-based policy and heuristics.
\method prunes BERT sequentially in a layer-wise manner.
For each layer, \method extracts features that are useful for the reinforcement learning agent to determine which attention head to be pruned from the current layer.
The final pruning policy found by the reinforcement learning agent is used to prune the corresponding layer.
Before \method proceeds to process the next layer, BERT is fine-tuned to recapture the information lost due to pruning attention heads.
An overview of \method transitioning from the second to the third layer of BERT is demonstrated in Figure \ref{fig:auber}.

Our contributions are summarized as follows:
\begin{itemize*}
\item \textbf{Regularization.}
BERT is prone to overfitting when the training dataset is too small.
\method effectively prunes appropriate attention heads to decrease the model capacity and regularizes BERT.
\item \textbf{Automation.}
By leveraging reinforcement learning, we automate the process of regularization of BERT. Instead of depending on hand-crafted policies or heuristics which often yield suboptimal results, \method inspects the current state of BERT and automatically chooses which attention head should be pruned.
\item \textbf{Experiments.}
We perform extensive experiments, and show that \method successfully regularizes BERT improving the performance metric by up to $10\%$ and outperforms other head pruning methods.
Through ablation study, we empirically show that our design choices for \method are effective.
\end{itemize*}

The rest of this paper is organized as follows.
Section~\ref{sec:prelim} explains preliminaries.
Section~\ref{sec:proposed} describes our proposed method, \method.
Section~\ref{sec:experim} presents experimental results.
After discussing related works in Section~\ref{sec:related},
we conclude in Section~\ref{sec:conclusion}. 

\section{Preliminary}
\label{sec:prelim}
We describe preliminaries on multi-headed self-attention (Section \ref{sec:prelim:self-att}), BERT (Section \ref{sec:prelim:BERT}), and deep Q-learning (Section \ref{sec:prelim:dqn}).

\subsection{Multi-Headed Self-Attention}
\label{sec:prelim:self-att}

A self-attention function maps a query vector and a set of key-value vector pairs to an output.
%
We compute the query, key, and value vectors by multiplying the input embeddings $Q, K, V\in \mathbb{R}^{N\times d}$ with the parametrized matrices $W^Q\in \mathbb{R}^{d\times n}$, $W^K\in \mathbb{R}^{d\times n}$, and $W^V\in \mathbb{R}^{d\times m}$ respectively,
where $N$ is the number of tokens in the sentence,
and $n, m$, and $d$ are query, value, and embedding dimension respectively.
In multi-headed self-attention, $H$ independently parameterized self-attention heads are applied in parallel to project the input embeddings into multiple representation subspaces.
Each attention head contains parameter matrices
$W_i^Q\in \mathbb{R}^{d\times n}$, $W_i^K\in \mathbb{R}^{d\times n}$, and $W_i^V\in \mathbb{R}^{d\times m}$.
Output matrices of $H$ independent self-attention heads are concatenated and once again projected by a matrix $W^O\in \mathbb{R}^{Hm\times d}$ to obtain the final result.
This process can be represented as:

\begin{equation}
	MultiHeadAtt (Q, K, V) = Concat( Att_{1...H}(Q,K,V) ) W^O
\end{equation}
where
\begin{align}
	Att_i(Q,K,V)=softmax(\dfrac{(QW_i^Q)(KW_i^K)^T}{\sqrt{n}})VW_i^V
\end{align}

\subsection{BERT}
\label{sec:prelim:BERT}
\vspace{-1mm}
BERT (\cite{BERT}) is a multi-layer Transformer (\cite{Transformer}) pre-trained on masked language model and next sentence prediction tasks.
It is then fine-tuned on specific tasks including question answering and language inference.
It achieved state-of-the-art performance on a variety of downstream natural language processing tasks.
BERT-base has 12 Transformer blocks and each block has 12 self-attention heads.
Despite its success in various natural language processing tasks, BERT sometimes overfits when the training dataset is small due to over-parameterization: $110$M parameters for BERT-base.
Thus, there has been a growing interest in BERT regularization through various methods such as pruning or dropout (\cite{MixOut}).

\subsection{Deep Q-learning}
\label{sec:prelim:dqn}
\vspace{-1mm}
A deep Q network (DQN) is a multi-layered neural policy network that outputs a vector of action-value pairs for a given state $s$.
For a $d_s$-dimensional state space and an action space containing $d_a$ actions, the neural network is a function from $\mathbb{R}^{d_s}$ to $\mathbb{R}^{d_a}$.
Two important aspects of the DQN algorithm as proposed by \cite{DQN} are the use of a target network, and the use of experience replay.
The target network is the same as the policy network except that its parameters are copied every $\tau$ steps from the policy network.
For the experience replay, observed transitions are stored for some time and sampled uniformly from this memory bank to update the network.
Both the target network and the experience replay dramatically improve the performance of the algorithm.

\section{Proposed Method}
\label{sec:proposed}
\vspace{-1mm}
We propose \method, our method for regularizing BERT
by automatically learning to prune attention heads from BERT.
After presenting the overview of the proposed method in Section \ref{sec:proposed:overview},
we describe how we frame the problem of pruning attention heads into a reinforcement learning problem in Section \ref{sec:proposed:RL}.
Then, we propose \method in Section \ref{sec:proposed:AUBER}.
\vspace{-1mm}

\subsection{Overview}
\label{sec:proposed:overview}
\vspace{-2mm}
We observe that BERT is prone to overfitting for tasks with a few training data.
However, the existing head pruning methods rely on hand-crafted heuristics and hyperparamters, which give sub-optimal results.
The goal of \method is to automate the pruning process for successful regularization.
Designing such regularization method entails the following challenges:
\begin{enumerate}
	\item \textbf{Automation.} How can we automate the head pruning process for regularization without resorting to sub-optimal heuristics?
	\item \textbf{State representation.}
When automating the regularization process as a reinforcement learning problem, how can we represent the state of BERT in a way useful for the pruning?
	\item \textbf{Action search space scalability.}
BERT has many parameters, many layers, and many attention heads in each layer.
When automating the regularization process of BERT as a reinforcement learning problem,
how can we handle prohibitively large action search space for pruning?
\end{enumerate}
We propose the following main ideas to address the challenges:
\begin{enumerate}
	\item \textbf{Reinforcement learning.} 
		We exploit reinforcement learning, specifically DQN, with accuracy enhancement as reward.
		\blue{It is natural to leverage DQN for these problems that have discrete action space (\cite{SuttonBook}).}
		Experience replay also allows efficient usage of previous experiences and stable convergence (\cite{DQN}).
	\item \textbf{L1 norm of value matrix.} We use L1 norm of value matrix of each attention head to represent initial state of a layer. When a head is pruned, the corresponding value is set to 0.
	\item \blue{\textbf{Dually-greedy manner.} We prune the attention heads layer by layer sequentially to reduce the search space.
		Moreover, we prune one attention head at one time instead of handling all possible pruning methods at once so that action search space becomes more scalable.}
\end{enumerate}

\subsection{Automated Regularization with Reinforcement Learning}
\label{sec:proposed:RL}

\blue{
\method leverages reinforcement learning for efficient search of regularization strategy without relying on heuristics.
We exploit DQN among various reinforcement learning frameworks to take advantage of experience replay and to easily handle discrete action space.
Here we introduce the detailed setting of reinforcement learning framework.
}

\textbf{Initial state}
\blue{
As mentioned in section \ref{sec:prelim:BERT}, layer $l$ has multiple attention heads, each of which has its own query, key, and value matrices.
For layer $l$ of BERT, we derive the initial state $s_l$ using L1 norm of the value matrix of each attention head.
Further details for this computation method is elaborated in section \ref{sec:proposed:state}.
}

\textbf{Action}
The action space $a$ of \method is discrete.
For a BERT model with $H$ attention heads per layer, the number of possible actions is $H+1$ (i.e. $a \in \{1, 2, \dots, H, H+1 \}$).
When the action $a=i \in \{1, 2, \dots, H-1, H \}$ is chosen, the corresponding $i^{th}$ attention head is pruned.
The action $a = H+1$ signals the DQN agent to quit pruning.
\blue{With a continuous action space (e.g. effective sparsity ratio), a separate heuristic-based pruning algorithm must be used in order to choose which attention heads should be pruned.
However, having a discrete action space allows the reinforcement learning agent to automatically infer the expected reward for each possible pruning policy, thereby minimizing the usage of error-prone heuristics.}

\textbf{Next State}
After the $i^{th}$ head is pruned, the value of $i^{th}$ index of $s_l$ is set to 0.
This modified state is provided as the next state to the agent.
When the action $a=H+1$, the next state is set to $None$.
\blue{This mechanism allows the agent to recognize which attention heads have been pruned and decide the next best pruning policy based on past decisions.}

\textbf{Reward}
The reward of \method is the change in accuracy

\vspace{-3mm}
\begin{align}
\Delta acc = current\_accuracy - prev\_accuracy
\end{align}

where $current\_accuracy$ is the accuracy of the current BERT model evaluated on a validation set,
and $prev\_accuracy$ is the accuracy obtained from the previous state
or the accuracy of the original BERT model if no attention heads are pruned.
If we set the reward simply as $current\_accuracy$, DQN cannot capture the differences among reward values if the changes in accuracy are relatively small.
Setting the reward as the change in accuracy has the normalization effect, thus stabilizing the training process of the DQN agent.
The reward for action $a = H + 1$ is a hyper-parameter that can be adjusted to encourage or discourage active pruning. In \method, it is set to $0$ to encourage the DQN agent to prune only when the expected change in accuracy is positive.

\textbf{Fine-tuning}
After the best pruning policy for layer $l$ of BERT is found, the BERT model pruned according the best pruning policy is fine-tuned with a smaller learning rate.
\blue{This fine-tuning step is crucial since it adjusts the weights of remaining attention heads to compensate for information lost due to pruning}.
Then, the initial state of layer $l+1$ is calculated and provided to the agent.
Since frequent fine-tuning may lead to overfitting, we separate the training dataset into two: a mini-validation dataset and a mini-training dataset.
The mini-validation dataset is the dataset on which the pruned BERT model is evaluated on to return a reward.
After the optimal pruning policy is determined by using the mini-validation dataset,
the mini-training dataset is used to fine-tune the pruned model.
When all layers are pruned by \method, the final model is fine-tuned with the entire training dataset with early stopping.

\blue{
\subsection{State Representation}
\label{sec:proposed:state}

The initial state $s_l$ of layer $l$ of BERT is computed through following procedure.
We first calculate the L1 norm of the value matrix of each attention head.
Then, we standardize the norm values to have a mean $\mu=0$ and a standard deviation $\sigma=1$.
Finally, the $softmax$ function is applied to the norm values to yield $s_l$.
We devise the method based on the following lemma.

\begin{lemma}
\label{initstatelemma}
For a layer with $H$ heads,
let $N$ be the number of tokens in the sentence
and
$m$, $n$, and $d$ be the value, query, and embedding dimension respectively.
Let $Q, K, V \in \mathbb{R}^{N\times d}$ be the input query, key, and value matrices, and $W_i^Q$, $W_i^K$, and $W_i^V$ be the weight parameters of the $i^{th}$ head such that $W_i^Q, W_i^K\in \mathbb{R}^{d\times n}$ and $W_i^V\in \mathbb{R}^{d\times m}$.
Let $O_i$ be the output of the $i^{th}$ head.
Then, $\lVert O_i \rVert_1 \le C \lVert W_i^V \rVert_1$ for the constant $C=N\lVert V\rVert_1$.
\end{lemma}
\vspace{-3mm}
\begin{proof}
	See the appendix.
\end{proof}
\vspace{-3mm}
}
This lemma provides the theoretical insight that L1 norm of the value matrix of a head bounds the L1 norm of its output matrix, which implies the importance of the head in the layer.

\subsection{\method: Automated BERT Regularization}
\label{sec:proposed:AUBER}
\vspace{-1mm}
Our DQN agent processes the BERT model in a layer-wise manner.
For each layer $l$ with $H$ attention heads, the agent receives an initial layer embedding $s_l$ which encodes useful characteristics of this layer.
Then, the agent outputs the index of an attention head that is expected to increase or maintain the training accuracy when removed.
After an attention head $i$ is pruned, the value of the $i^{th}$ index of $s_l$ is set to 0, and it is provided as the next state to the agent.
This process is repeated until the action $a = H+1$.
The model pruned up to layer $l$ is fine-tuned on the training dataset, and a new initial layer embedding $s_{l+1}$ is calculated from the fine-tuned model.

Algorithm \ref{alg:AUBER_alg} illustrates the process of \method.

\begin{algorithm}[H]
\label{alg:AUBER_alg}
\SetAlgoLined
	\SetKwFunction{isOddNumber}{isOddNumber}
	\SetKwInOut{KwIn}{Input}
	\SetKwInOut{KwOut}{Output}
	\KwIn{A BERT model $B_t$ fine-tuned on task $t$.}
	\KwOut{Regularized $B_t$.}

	$L \leftarrow \#\ of\ layers\ in\ B_t$
	
	$E \leftarrow total\ \#\ of\ episodes$
	
	$H \leftarrow \#\ of\ attention\ heads\ per\ layer$
	
	\For{$l \leftarrow 0$ \KwTo $L - 1$}{
		Initialize policy net $P$ and target net $T$
		
		Initialize replay memory $M$
		
		$original\_accuracy \leftarrow eval(B_t)$
		
		\For{$e \leftarrow 0$ \KwTo $E - 1$}{
		
			$s_l \leftarrow state\ vector\ of\ layer\ l$
			
			$prune\_num \leftarrow 0$
			
			$action \leftarrow None$
			
			$prev\_accuracy \leftarrow original\_accuracy$
			
			\While{$action \neq H + 1$}{
				\uIf{$prune\_num = H - 1$}{
					$action \leftarrow H + 1$
					
					$s_l^* \leftarrow None$
					
					$reward \leftarrow 0$
					
				}
				\Else{
					$action \leftarrow P(s_l)$
					
					$B_t.prune\_head(action)$
					
					$prune\_num \leftarrow prune\_num + 1$
					
					$s_l[action] \leftarrow 0$
					
					$s_l^* \leftarrow s_l$
					
					$current\_accuracy \leftarrow eval(B_t)$
					
					$reward \leftarrow current\_accuracy - prev\_accuracy$
					
					$prev\_accuracy \leftarrow current\_accuracy$
					
				}
				
				$M.update(s_l, action, s_l^*, reward)$
				
			}
			
			$P.optimize(M,T)$
		
		}
		
		$B_t.finetune(t)$
		
    }

\caption{\method}
\end{algorithm}

\section{Experiments}
\label{sec:experim}
\vspace{-2mm}
We conduct experiments to answer the following questions of \method.
\vspace{-1mm}
\begin{itemize*}
\item \textbf{Q1. Accuracy (Section \ref{subsec:acc}).} Given a BERT model fine-tuned on a specific natural language processing task, how well does \method improve the performance of the model? 
\item \textbf{Q2. Ablation Study (Section \ref{subsec:ablation}).} 
	How useful is the \emph{L1 norm of the value matrices} of attention heads in representing the state of BERT? 
	How does the order in which the layers are processed by \method affect regularization?
\end{itemize*}

\vspace{-1mm}
\subsection{Experimental Setup}
\label{subsec: exp_setup}
\vspace{-1mm}
\textbf{Datasets.}
We perform downstream natural language processing tasks on four GLUE datasets - MRPC, CoLA, RTE, and WNLI.
We test \method on datasets that contain less that 10,000 training instances since past experiments report that fine-tuning sometimes fails when a target dataset has fewer than 10,000 training instances (\cite{BERT,STILTs}).
Detailed information of these datasets is described in Table~\ref{tab:data}.

\vspace{-3mm}
\
\begin{table}[h]
\caption{Datasets.}
\centering
\label{tab:data}
\resizebox{0.47\textwidth}{!}{
\begin{tabular}{lrrrr}
\toprule
\textbf{dataset} & \textbf{\# of classes} & \textbf{\# of train} & \textbf{\# of dev} \\
\midrule
MRPC\tablefootnote{\url{https://www.microsoft.com/en-us/download/details.aspx?id=52398}\label{r}}  & $2$ & $3668$ & $408$ \\
CoLA\tablefootnote{\url{https://nyu-mll.github.io/CoLA/}} & $2$ &  $8551$ & $1043$ \\
RTE\tablefootnote{\url{https://aclweb.org/aclwiki/Recognizing_Textual_Entailment}} & $2$ & $2490$ & $277$ \\
WNLI\tablefootnote{\url{https://cs.nyu.edu/faculty/davise/papers/WinogradSchemas/WS.html}} & $2$ & $635$ & $71$ \\
\bottomrule
\end{tabular}
}
\end{table}

\textbf{BERT Model.}
We use the pre-trained \textit{bert-base-cased} model provided by huggingface\footnote{\url{https://github.com/huggingface/transformers}}.
We fine-tune this model on each dataset mentioned in Table \ref{tab:data} to obtain the initial model.
Initial models for MRPC, CoLA, and WNLI are fine-tuned on the corresponding dataset for $3$ epochs, and the initial model for RTE is fine-tuned for $4$ epochs.
The max sequence length is set to $128$, the batch size per gpu is set to $32$.
The learning rate for fine-tuning initial models for MRPC, CoLA, and WNLI is set to $0.00002$, and the learning rate for fine-tuning the initial model for RTE is set to $0.00001$.

\textbf{Reinforcement Learning.}
We use a $4$-layer feedforward neural network for the DQN agent.
The dimension of input, output, and all hidden layers are set to $12$, $13$, and $512$ respectively.
LeakyReLU is applied after all layers except for the last one.
We use the epsilon greedy strategy for choosing actions.
The initial and final epsilon values are set to $1$ and $0.05$ respectively, and the epsilon decay value is set to $256$.
The replay memory size is set to $5000$, and the batch size for training the DQN agent is set to $128$.
The discount value $\gamma$ for the DQN agent is set to $1$.
The learning rate is set to $0.000002$ when fine-tuning BERT after processing a layer.
Before processing each layer, the training dataset is randomly split into $1:2$ to yield a mini-training dataset and a mini-validation dataset.
When fine-tuning the final mode, the patience value of early stopping is set to $20$.

\textbf{Competitors.}
We compare \method with other methods that prune BERT's attention heads.
As a simple baseline, we examine random pruning policy and note the method as Random. 
We examine two different pruning methods based on the importance score.
In both methods, if \method prunes $P$ number of attention heads from BERT, we also prune $P$ attention heads with the smallest importance scores to obtain the competitor model.
We denote the pruning method using the confidence score as Confidence.
The confidence score of an attention head is the average of the maximum attention weight; a high confidence score indicates that the weight is concentrated on a single token.
On the other hand, \cite{SixteenHeads} performs a forward and backward pass to calculate gradients and uses them to assign an importance score to each attention head.
\cite{SpecializedHeads} constructs a new loss function that minimizes classification error and the number of being used heads so that unproductive heads are pruned while maintaining the model performance.
We prune the same number of heads as \method by tuning hyperparameters for fair comparison.

\textbf{Implementation.}
We construct all models using PyTorch framework.
All the models are trained and tested on GeForce GTX 1080 Ti GPU.

\subsection{Accuracy}
\vspace{-1mm}
\label{subsec:acc}
We evaluate the performance of \method against competitors.
Table \ref{tab:perf_AUBER} shows the results on four GLUE datasets specified on Table \ref{tab:data}.
Note that \method outperforms its competitors on regularizing BERT that is fine-tuned on MRPC, CoLA, RTE, or WNLI.
While most of its competitors fail to improve performance of BERT on the dev dataset of MRPC and CoLA, \method improves the performance of BERT by up to $4\%$.

\begin{table}[t]
\centering
\caption{
Performance of AUBER and competitors on 4 GLUE tasks.
The performance after fine-tuning is measured.
\method gives the best performance for the same number of pruned heads.
Bold font and underlined font indicate the best and second best accuracy among competing pruning methods respectively.
}
\label{tab:perf_AUBER}
\resizebox{0.49 \textwidth}{!}{
\subtable[MRPC (Accuracy)]{
\begin{tabular}{lcc}
\toprule
Policy	& \# of heads pruned & dev \\
\midrule
Original	& - & 84.07 \\
\midrule
\method		& \multirow{5}{*}{\begin{tabular}[c]{@{}c@{}}20 \end{tabular}} & \textbf{86.03} \\
Random		& & 83.97 \\
Confidence & & 83.33 \\
\cite{SixteenHeads}	& & 83.09 \\
\cite{SpecializedHeads}	& & \underline{84.17} \\
\bottomrule
\end{tabular}
}
}
\resizebox{0.49 \textwidth}{!}{
\subtable[CoLA (Matthew's correlation)]{
\begin{tabular}{lcc}
\toprule
Policy	& \# of heads pruned & dev \\
\midrule
Original	& - & 57.01 \\
\midrule
\method		& \multirow{5}{*}{\begin{tabular}[c]{@{}c@{}}53 \end{tabular}} & \textbf{61.28} \\
Random		& & \underline{57.89} \\
Confidence & & 55.74 \\
\cite{SixteenHeads}	& & 54.73 \\
\cite{SpecializedHeads}	& & 57.24 \\
\bottomrule
\end{tabular}
}
}

\resizebox{0.49 \textwidth}{!}{
\subtable[RTE (Accuracy)]{
\begin{tabular}{lcc}
\toprule
Policy	& \# of heads pruned & dev \\
\midrule
Original	& - & 63.54 \\
\midrule
\method		& \multirow{5}{*}{\begin{tabular}[c]{@{}c@{}}87 \end{tabular}} & \textbf{66.43} \\
Random		& & 63.47 \\
Confidence & & \underline{64.26} \\
\cite{SixteenHeads}	& & 63.18 \\
\cite{SpecializedHeads}	& & 63.61 \\
\bottomrule
\end{tabular}
}
}
\resizebox{0.49 \textwidth}{!}{
\subtable[WNLI (Accuracy)]{
\begin{tabular}{lcc}
\toprule
Policy	& \# of heads pruned & dev \\
\midrule
Original	& - & 46.48 \\
\midrule
\method		& \multirow{5}{*}{\begin{tabular}[c]{@{}c@{}}86 \end{tabular}} & \textbf{56.34} \\
Random		& & 43.07 \\
Confidence & & \textbf{56.34} \\
\cite{SixteenHeads}	& & 54.93 \\
\cite{SpecializedHeads}	& & \textbf{56.34} \\
\bottomrule
\end{tabular}
}
}
\end{table}

\subsection{Ablation Study}
\vspace{-1mm}
\label{subsec:ablation}
Here we empirically demonstrate the effectiveness of our design choices for \method.
More specifically, we validate that the \emph{L1\ norm of value matrix} of each attention head effectively guides \method to predict the best action.
Moreover, we show that \method successfully regularizes BERT regardless of the direction in which the layers are processed.
Table \ref{tab:AUBER_state} shows the performances of the variants of \method on the four GLUE datasets listed on Table 1.

\vspace{-0.5mm}
\subsubsection{\method with the Key/Query Matrices as the State Vector}
\label{subsubsec:state}
\vspace{-1mm}
Among the query, key, and value matrices of each attention head, we show that the value matrix best represents the current state of BERT.
Here we evaluate the performance of \method against \method-Query and \method-Key.
\method-Query and \method-Key use the query and key matrices respectively to obtain the initial state.
Note that \method, which uses the value matrix to obtain state vectors, outperforms \method-Query and \method-Key on all four tasks.

\vspace{-0.5mm}
\subsubsection{\method with L2 norm of the Value Matrices as the State Vector}
\label{subsubsec:L2}
\vspace{-1mm}
L1 norm of the value matrices is used to compute the state vector based on the theoretical derivation.
In this ablation study, we experimentally show that the L1 norm of the value matrices is appropriate for state vector.
We set a new variant \method-L2 which leverages L2 norm of the value matrices to compute the initial state vector instead of L1 norm.
The performance of \method is far more superior than \method-L2 in most cases bolstering that L1 norm of the value matrices effectively represents the state of BERT.

\subsubsection{Effect of Processing Layers in a Different Order}
\label{subsubsec:opposite}
\vspace{-1mm}
We empirically demonstrate how the order in which the layers are processed affects the final performance.
We evaluate the performance of \method against \method-Reverse.
\method-Reverse processes BERT in an opposite direction (i.e. from $Layer 12$ to $Layer 1$ for BERT-base).
Note that both \method and \method-Reverse effectively regularize BERT, proving the effectiveness of \method regardless of the order in which BERT layers are pruned.
The differences in the final performance and the number of attention heads pruned can be attributed to the fine-tuning step after pruning each layer.
Since the fine-tuning step adjusts the weights of the remaining attention heads in order to take the previous pruning policies into account, processing BERT in different directions may lead to different adjustments in weights.
Varying updates on weights may make previously important attention head become unimportant and vice versa, thus resulting in different pruning policies and final accuracies.

\begin{table}[t]
\centering
\caption{
We compare \method with four of its variants: \method-Query, \method-Key, \method-L2, and \method-Reverse on four GLUE datasets to demonstrate the effectiveness of various ways to calculate the initial state.
\method-Query and \method-Key use the query and key matrices respectively, and \method-L2 leverages L2 norm of the value matrix to obtain the initial state.
\method-Reverse processes BERT starting from the final layer (e.g. $12^{th}$ layer for BERT-base).
Bold font indicates the best accuracy among competing pruning methods. 
}
\label{tab:AUBER_state}

\resizebox{0.49\textwidth}{!}{
\subtable[MRPC (Accuracy)]{
\begin{tabular}{lcc}
\toprule
Policy         				& \# of heads pruned 				& dev	\\
\midrule
\method           				& 20  	 		& 86.03		\\
\midrule
\method-Query 	& 		27	 		& 83.33	\\
\method-Key 	& 		42	 		& 84.56	\\
\midrule
\method-L2	&	33	&83.33 \\
\midrule
\method-Reverse	& 58		& \textbf{86.52} \\
\bottomrule
\end{tabular}}
}
\resizebox{0.49\textwidth}{!}{
\subtable[CoLA (Matthew's correlation)]{
\begin{tabular}{lccc}
\toprule
Policy         				& \# of heads pruned 				& dev	\\
\midrule
\method           				& 53  	 		& \textbf{61.28}		\\
\midrule
\method-Query 	& 		72	 		& 55.52	\\
\method-Key 	& 		55	 		& 55.78	\\
\midrule
\method-L2	&	63	& 54.85 \\
\midrule
\method-Reverse	& 57		& 59.48 \\
\bottomrule
\end{tabular}}
}

\resizebox{0.49\textwidth}{!}{
\subtable[RTE (Accuracy)]{
\begin{tabular}{lcc}
\toprule
Policy         				& \# of heads pruned 			& dev	\\
\midrule
\method           				& 87  	 		& \textbf{66.43}		\\
\midrule
\method-Query 	& 		83	 		& 65.34 	\\
\method-Key 	& 		86	 		& 63.18	\\
\midrule
\method-L2	& 61	& \textbf{66.43}	\\
\midrule
\method-Reverse	& 99		& 64.62 \\
\bottomrule
\end{tabular}}
}
\resizebox{0.49\textwidth}{!}{
\subtable[WNLI (Accuracy)]{
\begin{tabular}{lcc}
\toprule
Policy         				& \# of heads pruned 					& dev	\\
\midrule
\method           				&  86     		& \textbf{56.34}		\\
\midrule
\method-Query 	& 		96	 		& \textbf{56.34}	\\
\method-Key 	& 		101	 		& \textbf{56.34}	\\
\midrule
\method-L2	& 94	& 53.52 \\
\midrule
\method-Reverse	& 101	& 54.93 \\
\bottomrule
\end{tabular}}
}

\end{table}

\section{Related Work}
\label{sec:related}
\vspace{-1mm}
A number of studies focused on analyzing the effectiveness of multi-headed attention
(\cite{SpecializedHeads,SixteenHeads}).
These studies evaluate the importance of each attention head by measuring some heuristics such as the
average of its maximum attention weight,
where average is taken over tokens in a set of sentences used for evaluation,
or the expected sensitivity of the model to attention head pruning.
Their results show that a large percentage of attention heads with low importance scores can be pruned 
without significantly impacting performance.
However, they usually yield suboptimal results since they predetermine the order in which the attention heads are pruned by using heuristics. 

To prevent overfitting of BERT on downstream natural language processing tasks,
various regularization techniques are proposed.
A variant of dropout improves the stability of fine-tuning a big, 
pre-trained language model even with only a few training examples of a target task (\cite{MixOut}).
Other existing heuristics to prevent overfitting include choosing a small learning rate
or a triangular learning rate schedule, and a small number of iterations.

To automate the process of Convolutional Neural Network pruning, \cite{AMC} leverages reinforcement learning to determine the best sparsity ratio for each layer.
Important features that characterize a layer are encoded and provided to a reinforcement learning agent to determine how much of the current layer should be pruned.
To the best of our knowledge, \method is the first attempt to use reinforcement learning to prune attention heads from Transformer-based models such as BERT.

\section{Conclusion}
\label{sec:conclusion}
We propose \method, an effective method to regularize BERT by automatically pruning attention heads.
Instead of depending on heuristics or rule-based policies, \method leverages reinforcement learning to learn a pruning policy that determines which attention heads should be pruned for better regularization.
Experimental results demonstrate that \method effectively regularizes BERT, increasing the performance of the original model on the dev dataset by up to $10\%$.
In addition, we experimentally demonstrate the effectiveness of our design choices for \method.

\clearpage
\bibliography{iclr2021_conference}
\bibliographystyle{iclr2021_conference}

\newpage
\appendix
\section{Appendix}
\vspace{-1mm}
\addtocounter{lemma}{-1}
\subsection{Proof for Lemma 1}
\vspace{-1mm}
\begin{lemma}
\label{initstatelemma}
For a layer with $H$ heads,
let $N$ be the number of tokens in the sentence
and
$m$, $n$, and $d$ be the value, query, and embedding dimension respectively.
Let $Q, K, V \in \mathbb{R}^{N\times d}$ be the input query, key, and value matrices, and $W_i^Q$, $W_i^K$, and $W_i^V$ be the weight parameters of the $i^{th}$ head such that $W_i^Q, W_i^K\in \mathbb{R}^{d\times n}$ and $W_i^V\in \mathbb{R}^{d\times m}$.
Let $O_i$ be the output of the $i^{th}$ head.
Then, $\lVert O_i \rVert_1 \le C \lVert W_i^V \rVert_1$ for the constant $C=N\lVert V\rVert_1$.
\end{lemma}
\vspace{-1mm}
\begin{proof}
	For $i^{th}$ head in the layer, let
	\begin{equation}
		softmax_i = softmax(\dfrac{(QW_i^Q)(KW_i^K)^T}{\sqrt{n}} )
	\end{equation}
	and
	\begin{equation}
		v_i = VW_i^V.
	\end{equation}
	The output of the head, $O_i$, is evaluated as $O_i=softmax_iv_i$. Then,
	\begin{align}
		\lVert O_i \rVert _1 
		&= \sum_{j=1}^{T}{\sum_{k=1}^{m}{\lvert (O_i)_{jk}\rvert}} \\
		&= \sum_{j=1}^{T}{\sum_{k=1}^{m}{|(softmax_i)_{j\boldsymbol{\cdot}} \boldsymbol{\cdot} (v_i)_{\boldsymbol{\cdot} k}|}} \\
		&\le  \sum_{j=1}^{T}{\sum_{k=1}^{m}{\lVert (softmax_i)_{j\boldsymbol{\cdot}} \rVert_2 \lVert (v_i)_{\boldsymbol{\cdot} k} \rVert_2}} \\
		&= \sum_{j=1}^{T}{\lVert (softmax_i)_{j\boldsymbol{\cdot}} \rVert_2} \sum_{k=1}^{m}{\lVert (v_i)_{\boldsymbol{\cdot}k} \rVert_2}
	\end{align}
	Since the L1 norm of a vector is always greater than or equal to the L2 norm of the vector,
	\begin{align}
		\lVert O_i \rVert_1 
		&\le \sum_{j=1}^{T}{\lVert (softmax_i)_{j\boldsymbol{\cdot}} \rVert_1} \sum_{k=1}^{m}{\lVert (v_i)_{\boldsymbol{\cdot}k} \rVert_1} \\
		&\le T\sum_{k=1}^{m}{\lVert (v_i)_{\boldsymbol{\cdot}k}}\rVert_1 \\
		&= T\sum_{j=1}^{T}{\sum_{k=1}^{m}{\lvert (v_i)_{jk} \rvert}} \\
		&= T\sum_{j=1}^{T}{\sum_{k=1}^{m}{\lvert V_{j\boldsymbol{\cdot}} \boldsymbol{\cdot} (W_i^V)_{\boldsymbol{\cdot}k} \rvert}} \\
		&\le T\sum_{j=1}^{T}{\sum_{k=1}^{m}{\lVert V_{j\boldsymbol{\cdot}} \rVert_2 \lVert (W_i^V)_{\boldsymbol{\cdot}k} \rVert_2}} \\
		&= T\sum_{j=1}^{T}{\lVert V_{j\boldsymbol{\cdot}} \rVert_2 } \sum_{k=1}^{m}{\lVert (W_i^V)_{\boldsymbol{\cdot}k} \rVert_2} \\
		&\le  T\sum_{j=1}^{T}{\lVert V_{j\boldsymbol{\cdot}}\rVert_1} \sum_{k=1}^{m}{\lVert(W_i^V)_{\boldsymbol{\cdot}k}\rVert_1} \\
		&= T \lVert V \rVert_1 \lVert W_i^V \rVert_1
	\end{align}
	where the norm of the matrices is entrywise norm, $\lVert A \rVert_1 = \sum_j{\sum_k{A_{jk}}}$.
	All heads in the same layer take the same $V$ as input and $T$ is constant. Thus,
	\begin{equation}
		\lVert O^i \rVert_1 \le C\lVert W_i^V \rVert_1
	\end{equation}
	for the constant $C=T\lVert V\rVert_1$.
\end{proof}

\end{document}